\documentclass[12pt,english]{article}
\usepackage[T1]{fontenc}
\usepackage[latin9]{inputenc}
\usepackage{geometry}
\usepackage{color}
\usepackage{babel}
\usepackage{float}
\usepackage{amsmath}
\usepackage{graphicx}
\usepackage[authoryear]{natbib}
\usepackage[unicode=true,
 bookmarks=false,
 breaklinks=false,pdfborder={0 0 1},backref=false,colorlinks=true]
 {hyperref}
\hypersetup{citecolor=blue}

\makeatletter

\usepackage{cite}
\usepackage{indentfirst}
\usepackage{algorithm}
\usepackage{algorithmic}
\usepackage{caption}

\DeclareMathOperator{\tr}{tr}

\makeatother

\begin{document}

\title{The information bottleneck and geometric clustering}

\author{DJ Strouse\thanks{Department of Physics, Princeton University}~ \& David J. Schwab\thanks{Initiative for the Theoretical Sciences, CUNY Graduate Center}}
\maketitle
\begin{abstract}
The information bottleneck (IB) approach to clustering takes a joint distribution $P\!\left(X,Y\right)$ and maps the data $X$ to cluster labels $T$ which retain maximal information about $Y$ \citep{tishby1999ib}. This objective results in an algorithm that clusters data points based upon the similarity of their conditional distributions $P\!\left(Y\mid X\right)$. This is in contrast to classic ``geometric clustering'' algorithms such as $k$-means and gaussian mixture models (GMMs) which take a set of observed data points $\left\{ \mathbf{x}_{i}\right\} _{i=1:N}$ and cluster them based upon their geometric (typically Euclidean) distance from one another. Here, we show how to use the deterministic information bottleneck (DIB) \citep{strouse2017dib}, a variant of IB, to perform geometric clustering, by choosing cluster labels that preserve information about data point location on a smoothed dataset. We also introduce a novel method to choose the number of clusters, based on identifying solutions where the tradeoff between number of clusters used and spatial information preserved is strongest. We apply this approach to a variety of simple clustering problems, showing that DIB with our model selection procedure recovers the generative cluster labels. We also show that, in particular limits of our model parameters, clustering with DIB and IB is equivalent to $k$-means and EM fitting of a GMM with hard and soft assignments, respectively. Thus, clustering with (D)IB generalizes and provides an information-theoretic perspective on these classic algorithms.
\end{abstract}

\section{Introduction}

Unsupervised learning is a crucial component of building intelligent systems \citep{lecun2016nipstalk}, since such systems need to be able to leverage experience to improve performance even in the absence of feedback. One aspect of doing so is discovering discrete structure in data, a problem known as clustering \citep{mackay2002textbook}. In the typical setup, one is handed a set of data points $\left\{ \mathbf{x}_{i}\right\} _{i=1}^{N}$, and asked to return a mapping from data point label $i$ to a finite set of cluster labels $c$. The most basic approaches include $k$-means and gaussian mixture models (GMMs). GMMs cluster data based on maximum likelihood fitting of a probabilistic generative model. $k$-means can either be thought of as directly clustering data based on geometric (often Euclidean) distances between data points, or as a special case of GMMs with the assumptions of evenly sampled, symmetric, equal variance components.

The information bottleneck (IB) is an information-theoretic approach to clustering data $X$ that optimizes cluster labels $T$ to preserve information about a third ``target variable'' of interest $Y$. The resulting (soft) clustering groups data points based on the similarity in their conditional distributions over the target variable through the KL divergence, $\text{KL}\!\left[p\!\left(y\mid x_{i}\right)\mid p\!\left(y\mid x_{j}\right)\right]$. An IB clustering problem is fully specified by the joint distribution $P\!\left(X,Y\right)$ and the tradeoff parameter $\beta$ quantifying the relative preference for fewer clusters and more informative ones.

At first glance, it is not obvious how to use this approach to cluster geometric data, where the input is a set of data point locations $\left\{ \mathbf{x}_{i}\right\} _{i=1}^{N}$. For example, what is the target variable $Y$ that our clusters should retain information about? What should $P\!\left(X,Y\right)$ be? And how should one choose the tradeoff parameter $\beta$?

\citet{still2004geoIB} were the first to attempt to do geometric clustering with IB, and claimed an equivalence (in the large data limit) between IB and $k$-means. Unfortunately, while much of their approach is correct, it contained errors that nullify the main results. In the next section, we describe those errors and how to correct them. Essentially, their approach did not properly translate geometric information into a form that could be used correctly by an information-theoretic algorithm.

In addition to fixing this issue, we also choose to use a recently introduced variant of the information bottleneck called the deterministic information bottleneck (DIB) \citep{strouse2017dib}. We make this choice due to the different way in which IB and DIB use the number of clusters provided to them. IB is known to use all of the clusters it has access to, and thus clustering with IB requires a search both over the number of clusters $N_{c}$ as well as the the parsimony-informativeness tradeoff parameter $\beta$ \citep{slonim2005clusteringIB}. DIB on the other hand has a built-in preference for using as few clusters as it can, and thus only requires a parameter search over $\beta$. Moreover, DIB's ability to select the number of clusters to use for a given $\beta$ leads to a intuitive model selection heuristic based on the robustness of a clustering result across $\beta$ that we show can recover the generative number of clusters in many cases. 

In the next section, we more formally define the geometric clustering problem, the IB approach of \citet{still2004geoIB}, and our own DIB approach. In section~\ref{sec:Results:-geometric-clustering}, we show that our DIB approach to geometric clustering behaves intuitively and is able to recover the generative number of clusters with only a single free parameter (the data smoothing scale $s$). In section~\ref{sec:GMMs}, we discuss the relationship between our approach and $k$-means/GMMs, showing that in particular limits, clustering with DIB and IB is equivalent to $k$-means/EM fitting of a GMM with hard and soft assignments, respectively. Our approach thus provides a novel information-theoretic approach to geometric clustering, as well as an information-theoretic perspective on these classic clustering methods.

\section{Geometric clustering with the (deterministic) information bottleneck \label{sec:clustering-with-DIB}}

In a geometric clustering problem, we are given a set of $N$ observed data points $\left\{ \mathbf{x}_{i}\right\} _{i=1:N}$ and asked to provide a weighting $q\!\left(c\mid i\right)$ that categorizes data points into (possibly multiple) clusters such that data points ``near'' one another are in the same cluster. The definition of ``near'' varies by algorithm: for $k$-means, for example, points in a cluster are closer to their own cluster mean than to any other cluster mean.

In an information bottleneck (IB) problem, we are given a joint distribution $P\!\left(X,Y\right)$ and asked to provide a mapping $q\!\left(t\mid x\right)$ such that $T$ contains the ``relevant'' information in $X$ for predicting $Y$. This goal is embodied by the information-theoretic optimization problem

\begin{align}
q_{\text{IB}}^{*}\!\left(t\mid x\right) & =\text{\ensuremath{\underset{q\left(t\mid x\right)}{\operatorname{argmin}}}}\,\,I\!\left(X,T\right)-\beta I\!\left(T,Y\right),
\end{align}
subject to the Markov constraint $T\leftrightarrow X\leftrightarrow Y$. $\beta$ is a free parameter that allows for setting the desired balance between the compression encouraged by the first term and the relevance encouraged by the second; at small values, we throw away most of $X$ in favor of a succinct representation for $T$, while for large values of $\beta$, we retain nearly all the information that $X$ has about $Y$.

This approach of squeezing information through a latent variable bottleneck might remind some readers of a variational autoencoder (VAE) \citep{kingma2014vae}, and indeed IB has a close relationship with VAEs. As pointed out in \citep{alemi2017vib}, a variational version of IB can essentially be seen as the supervised generalization of a VAE, which is typically an unsupervised algorithm.

We are interested in performing geometric clustering with the information bottleneck. For the purposes of this paper, we will focus on a recent alternative formulation of the IB, called the deterministic information bottleneck (DIB) \citep{strouse2017dib}. We do this because the DIB's cost function more directly encourages the use of as few clusters as possible, so initialized with $n_{c}^{\text{max}}$ clusters, it will typically converge to a solution with far fewer. Thus, it has a form of model selection built in that will prove useful for geometric clustering \citep{strouse2017dib}. IB, on the other hand, will tend to use all $n_{c}^{\text{max}}$ clusters, and thus requires an additional search over this parameter \citep{slonim2005clusteringIB}. DIB also differs from IB in that it leads to a hard clustering instead of a soft clustering.

Formally, the DIB setup is identical to that of IB except that the mutual information term $I\!\left(X;T\right)$ in the cost functional is replaced with the entropy $H\!\left(T\right)$

\begin{align}
q_{\text{DIB}}^{*}\!\left(t\mid x\right) & =\text{\ensuremath{\underset{q\left(t\mid x\right)}{\operatorname{argmin}}}}\,\,H\!\left(T\right)-\beta I\!\left(T,Y\right).\label{eq:DIB_cost}
\end{align}
This change to the cost functional leads to a hard clustering with the form \citep{strouse2017dib}

\begin{align}
q_{\text{DIB}}^{*}\!\left(t\mid x\right) & =\delta\!\left(t-t^{*}\!\left(x\right)\right)\\
t^{*} & =\text{\ensuremath{\underset{t}{\operatorname{argmax}}}}\,\,\log q\!\left(t\right)-\beta \text{KL}\!\left[p\!\left(y\mid x\right)\mid q\!\left(y\mid t\right)\right]\\
q\!\left(t\right) & =\sum_{x}q\!\left(t\mid x\right)p\!\left(x\right)\\
q\!\left(y\mid t\right) & =\frac{1}{q\!\left(t\right)}\sum_{x}q\!\left(t\mid x\right)p\!\left(x\right)p\!\left(y\mid x\right),
\end{align}
where the above equations are to be iterated to convergence from some initialization. The IB solution \citep{tishby1999ib} simply replaces the first two equations with
\begin{align}
q_{\text{IB}}^{*}\!\left(t\mid x\right) & =\frac{q\!\left(t\right)}{Z\!\left(x,\beta\right)}\exp\!\left[-\beta \text{KL}\!\left[p\!\left(y\mid x\right)\mid q\!\left(y\mid t\right)\right]\right],\label{eqn:IBenc}
\end{align}
which can be seen as replacing the argmax in DIB with a soft max.

\sloppy The (D)IB is referred to as a ``distributional clustering'' algorithm \citep{slonim2001docIB} due to the KL divergence term $d\!\left(x,t\right)=\text{KL}\!\left[p\!\left(y\mid x\right)\mid q\!\left(y\mid t\right)\right]$, which can be seen as measuring how similar the data point conditional distribution $p\!\left(y\mid x\right)$ is to the cluster conditional, or mixture of data point conditionals, $q\!\left(y\mid t\right)=\sum_{x}q\!\left(x\mid t\right)p\!\left(y\mid x\right)$. That is, a candidate point $x^{'}$ will be assigned to a cluster based upon how similar its conditional $p\!\left(y\mid x^{'}\right)$ is to the conditionals $p\!\left(y\mid x\right)$ for the data points $x$ that make up that cluster. Thus, both DIB and IB cluster data points based upon the conditionals $p\!\left(y\mid x\right)$.

To apply (D)IB to a geometric clustering problem, we must choose how to map the geometric clustering dataset $\left\{ \mathbf{x}_{i}\right\} _{i=1:N}$ to an appropriate IB dataset $P\!\left(X,Y\right)$. First, what should $X$ and $Y$ be? Since $X$ is the data being clustered by IB, we'll choose that to be the \emph{data point index} $i$. As for the target variable $Y$ that we wish to maintain information about, it seems reasonable to choose the \emph{data point location} $\mathbf{x}$ (though we will discuss alternative choices later). Thus, we want to cluster data indices $i$ into cluster indices $c$ in a way that maintains as much possible info about the location $\mathbf{x}$ as possible \citep{still2004geoIB}.

Now, how should we choose the joint distribution $p\!\left(i,\mathbf{x}\right)=p\!\left(\mathbf{x}\mid i\right)p\!\left(i\right)$? At first glance, one might choose $p\!\left(\mathbf{x}\mid i\right)=\delta_{\mathbf{x}\mathbf{x}_{i}}$, since data point $i$ was observed at location $\mathbf{x}_{i}$. The reason \emph{not} to do this lies with the fact that (D)IB is a distributional clustering algorithm, as discussed two paragraphs above. Data points are compared to one another through their conditionals $p\!\left(\mathbf{x}\mid i\right)$, and with the choice of a delta function, there will be no overlap unless two data points are on top of one another. That is, choosing $p\!\left(\mathbf{x}\mid i\right)=\delta_{\mathbf{x}\mathbf{x}_{i}}$ leads to a $\text{KL}$ divergence that is either infinite for data points at different locations, or zero for data points that lie exactly on top of one another, i.e. $\text{KL}\!\left[p\!\left(\text{\textbf{x}}\mid i\right)\mid p\!\left(\text{\textbf{x}}\mid j\right)\right]=\delta_{\mathbf{x}_{i}\mathbf{x}_{j}}$. Trivially, the resulting clustering would assign each data point to its own cluster, grouping only data points that are identical. Put another way, all relational information in an IB problem lies in the joint distribution $P\!\left(X,Y\right)$. If one wants to perform geometric clustering with an IB approach, then geometric information must somehow be injected into that joint distribution, and a series of delta functions does not do that. A previous attempt at linking IB and $k$-means made this mistake \citep{still2004geoIB}. Subsequent algebraic errors were tantamount to incorrectly introducing geometric information into IB, precisely in the way that such geometric information appears in $k$-means, and resulting in an algorithm that is not IB. We describe these errors in more detail in section~\ref{sec:Appendix:-errors}.

\begin{figure*}[t]
\begin{centering}
\includegraphics[scale=0.23]{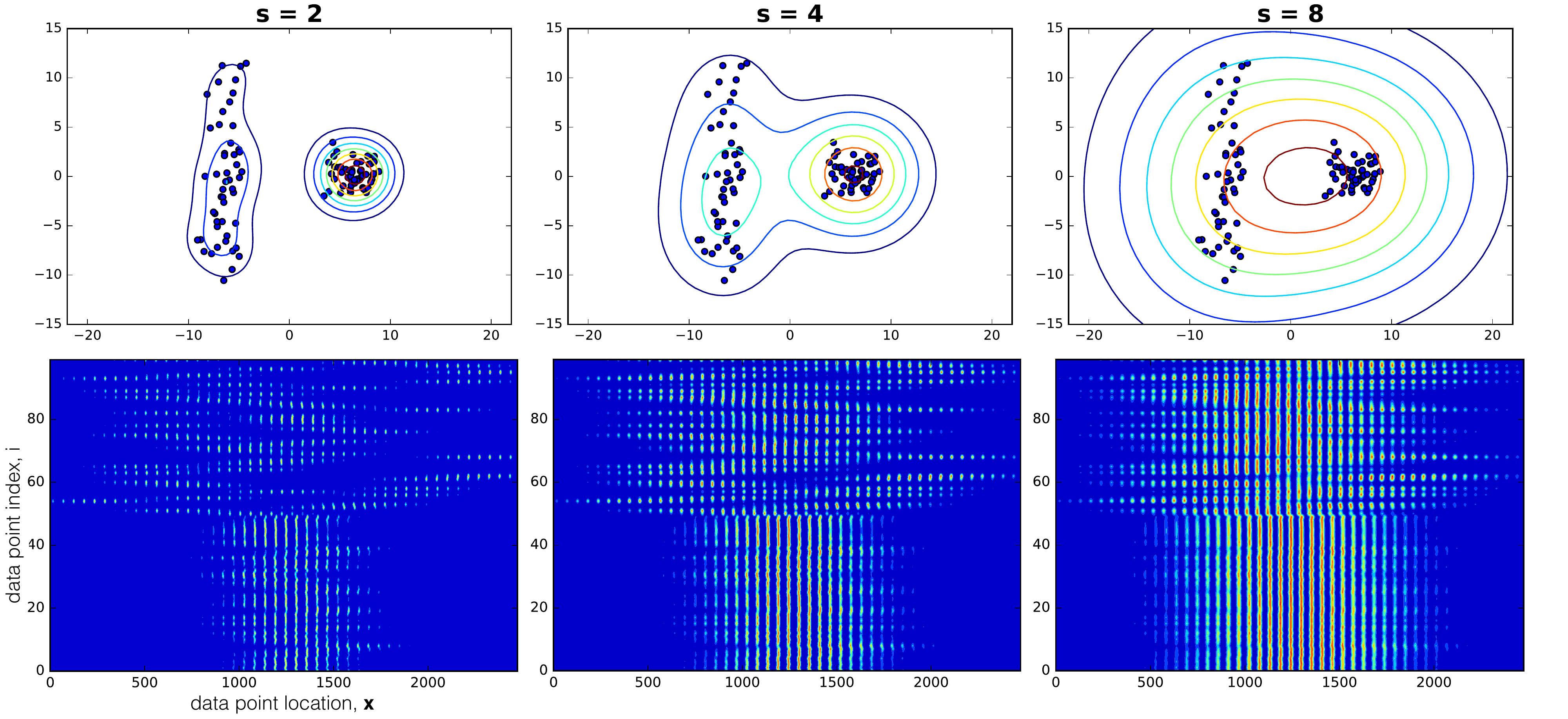}
\par\end{centering}
\caption{\textbf{Illustration of data smoothing procedure.}
\label{fig:smoothing}
Example dataset with one symmetric and one skew cluster. \emph{Top row}: scatterplot of data points with smoothed probability distribution overlaid. \emph{Bottom row}: heat map of the joint distribution $P\!\left(i,\mathbf{x}\right)$ that is fed into DIB. The two spatial dimensions in the top row are binned and concatenated into a single dimension (on the horizontal axis) in the bottom row, which is the source of the ``striations.''}
\end{figure*}

Based on the problems identified with using delta functions, a better choice for the conditionals is something spatially extended, such as:
\begin{align}
p\!\left(\mathbf{x}\mid i\right) & \propto\exp\!\left[-\frac{1}{2s^{2}}d\!\left(\mathbf{x},\mathbf{x}_{i}\right)\right],\label{eq:p(x|i)}
\end{align}
where $s$ sets the geometric scale or units of distance, and $d$ is a distance metric, such as the Euclidean distance $d\!\left(\mathbf{x},\mathbf{x}_{i}\right)=\left\Vert \mathbf{x}-\mathbf{x}_{i}\right\Vert ^{2}$. If we indeed use the Euclidean distance, then $p\!\left(\mathbf{x}\mid i\right)$ will be (symmetric) gaussian (with variance $s^{2}$), and this corresponds to gaussian smoothing our data. In any case, the obvious choice for the marginal is $p\!\left(i\right)=\frac{1}{N}$, where $N$ is the number of data points, unless one has a reason a priori to favor certain data points over others. These choices for $p\!\left(i\right)$ and $p\!\left(\mathbf{x}\mid i\right)$ determine completely our dataset $p\!\left(i,\mathbf{x}\right)=p\!\left(\mathbf{x}\mid i\right)p\!\left(i\right)$. Figure~\ref{fig:smoothing} contains an illustration of this data smoothing procedure. We will explore the effect of the choice of smoothing scale $s$ throughout this paper.

With the above choices, we have a fully specified DIB formulation of a geometric clustering problem. Using our above notational choices, the equations for the $n^{\text{th}}$ step in the iterative DIB solution is \citep{strouse2017dib}

\begin{align}
q^{\left(n\right)}\!\left(c\mid i\right) & =\delta\!\left(c-c^{*\left(n\right)}\!\left(i\right)\right)\label{eqn:q(c|i)}\\
c^{*\left(n\right)}\!\left(i\right) & \equiv\text{\ensuremath{\underset{c}{\operatorname{argmax}}}}\,\,\log q^{\left(n-1\right)}\!\left(c\right)-\beta \text{KL}\!\left[p\!\left(\mathbf{x}\mid i\right)\mid q^{\left(n-1\right)}\!\left(\mathbf{x}\mid c\right)\right]\label{eqn:c*}\\
q^{\left(n\right)}\!\left(c\right) & =\frac{n_{c}^{\left(n\right)}}{N}\label{eqn:q(c)}\\
q^{\left(n\right)}\!\left(\mathbf{x}\mid c\right) & =\sum_{i}q^{\left(n\right)}\!\left(i\mid c\right)p\!\left(\mathbf{x}\mid i\right) =\frac{1}{n_{c}^{\left(n\right)}}\sum_{i\in S_c^{\left(n\right)}}p\!\left(\mathbf{x}\mid i\right),\label{eqn:q(x|c)}
\end{align}
where $S_c^{\left(n\right)} \equiv \left\{ i: c^{* \left(n\right)} \left(i \right)=c \right\}$ is the set of indices of data points assigned to cluster $c$ at step $n$, and $n_{c}^{\left(n\right)}\equiv \left| S_c^{\left(n\right)}\right| $ is the number of data points assigned to cluster $c$ at step $n$. This process is summarized in algorithm~\ref{alg:DIBclustering}.

\begin{algorithm}[tb]
\caption{Geometric clustering with DIB.}
\begin{algorithmic}
\label{alg:DIBclustering}
\STATE\textbf{Input}: data $\left\{ \mathbf{x}_{i}\right\} _{i=1:N}$, tradeoff parameter $\beta$, smoothing scale $s$
\STATE Initialize cluster assignments $c^{*\left(0\right)}\!\left(i\right)$
\STATE Initialize cluster marginals $q^{\left(0\right)}\!\left(c\right)$ using eqn~\ref{eqn:q(c)}
\STATE Initialize cluster conditionals $q^{\left(0\right)}\!\left(\mathbf{x}\mid c\right)$ using eqn~\ref{eqn:q(x|c)}
\STATE Initialize step count $n=0$
\WHILE{not converged}
  \STATE $n \mathrel{+}= 1$
  \STATE Update cluster assignments $c^{*\left(n\right)}\!\left(i\right)$ using eqn~\ref{eqn:q(c|i)}
  \STATE Update cluster marginals $q^{\left(n\right)}\!\left(c\right)$ using eqn~\ref{eqn:q(c)}
  \STATE Update cluster conditionals $q^{\left(n\right)}\!\left(\mathbf{x}\mid c\right)$ using eqn~\ref{eqn:q(x|c)}
\ENDWHILE \end{algorithmic} 
\end{algorithm}

\begin{figure}[H]
\begin{centering}
\includegraphics[scale=0.16]{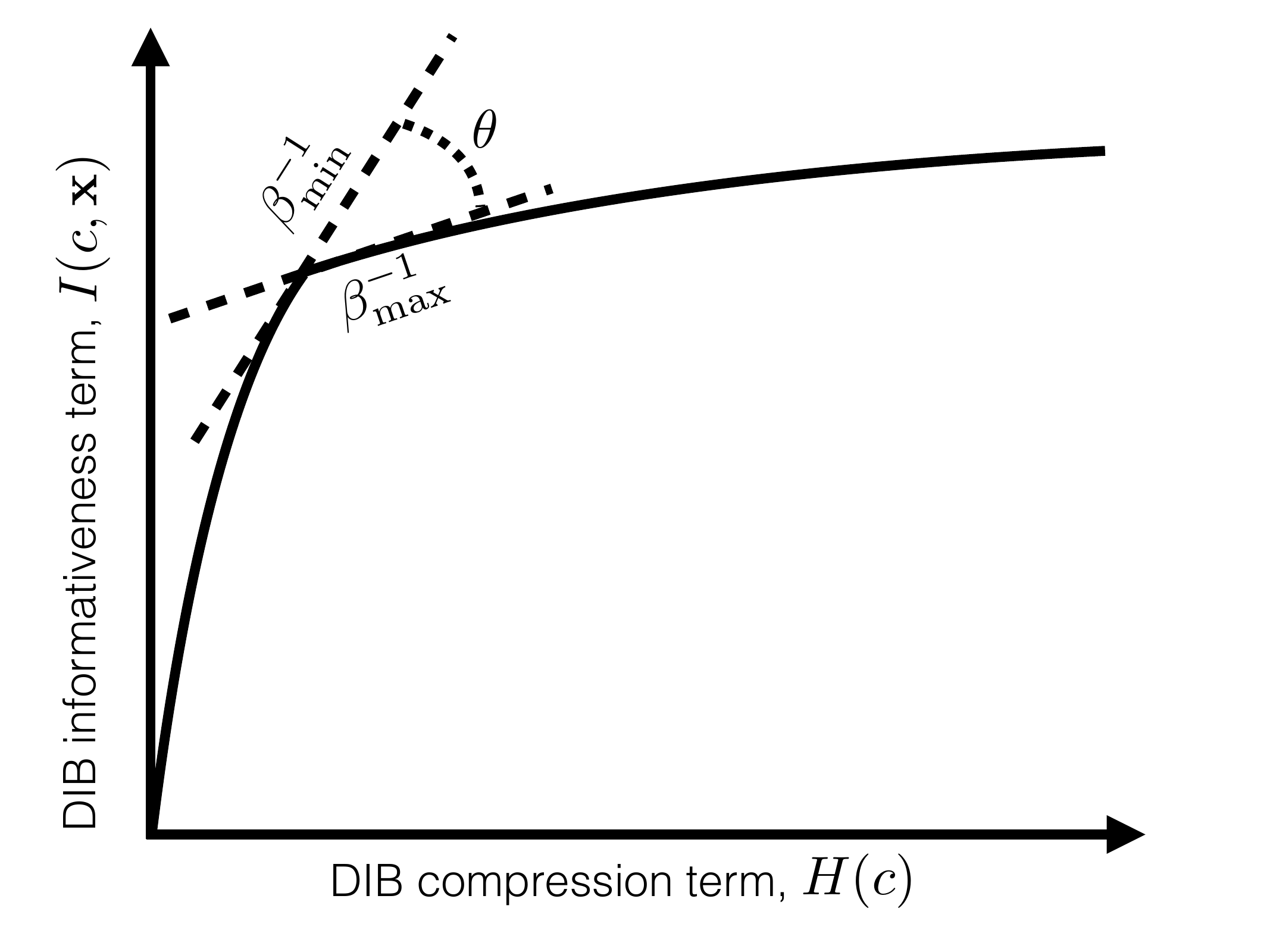}
\par\end{centering}
\caption{\textbf{``Kinks'' in DIB information curve as model selection}.\label{fig:kink}
$\beta_{\text{min}}$ and $\beta_{\text{max}}$ are the smallest and largest $\beta$ at which the solution at the kink is valid. Thus, $\beta_{\text{min}}^{-1}$ and $\beta_{\text{max}}^{-1}$ are the slopes of upper and lower dotted lines. The ``kink angle'' is then $\theta=\frac{\pi}{2}-\arctan\!\left(\beta_{\text{min}}\right)-\arctan\!\left(\beta_{\text{max}}^{-1}\right)$. It is a measure of how robust a solution is to the choice of $\beta$; thus high values of $\theta$ indicate solutions of particular interest.}
\end{figure}

Note that this solution contains $\beta$ as a free parameter. As discussed above, it allows us to set our preference between solutions with fewer clusters and those that retain more spatial information. It is common in the IB literature to run the algorithm for multiple values of $\beta$ and to plot the collection of solutions in the ``information plane'' with the relevance term $I\!\left(Y;T\right)$ on the $y$-axis and the compression term $I\!\left(X;T\right)$ on the $x$-axis \citep{palmer2015predictive,creutzig2009pastfuture,chechki2005gib,slonim2005clusteringIB,still2004finitedataIB,tishby2015deepIB,rubin2016auditoryIB,strouse2017dib,shwartzziv2017deepIB}. The natural such plane for the DIB is with the relevance term $I\!\left(Y;T\right)$ on the $y$-axis and \emph{its} compression term $H\!\left(T\right)$ on the $x$-axis \citep{strouse2017dib}. The curve drawn out by (D)IB solutions in the information plane can be viewed as a Pareto-optimal boundary of how much relevant information can be extracted about $Y$ given a fixed amount of information about $X$ (IB) or representational capacity by $T$ (DIB) \citep{strouse2017dib}. Solutions lying below this curve are of course suboptimal, but a priori, the (D)IB formalism doesn't tell us how to select a single solution from the family of solutions lying on the (D)IB boundary. Intuitively however, when faced with a boundary of Pareto-optimality, if we must pick one solution, its best to choose one at the ``knee'' of the curve. Quantitatively, the ``knee'' of the curve is the point where the curve has its maximum magnitude second derivative. In the most extreme case, the second derivative is infinite when there is a ``kink'' in the curve, and thus the largest kinks might correspond to solutions of particular interest. In our case, since the slope of the (D)IB curve at any given solution is $\beta^{-1}$ (which can be read off from the cost functionals), kinks also indicate solutions that are valid over a wide range of $\beta$. So large kinks additionally correspond to solutions robust to model hyperparameters, in the sense that they optimize a wide range of (D)IB tradeoffs. Such robust solutions should correspond to ''real'' structure in the data. Quantitatively, we can measure the size of a kink by the angle $\theta$ of the discontinuity it causes in the slope of the curve; see figure~\ref{fig:kink} for details. We will show in the next section that searches for solutions with large $\theta$ result in recovering the generative cluster labels for geometric data, including the correct number of clusters.

Note that this model selection procedure would not be possible if we had chosen to use IB instead of DIB. IB uses all the clusters available to it, regardless of the choice of $\beta$. Thus, all solutions on the curve would have the same number of clusters anyway, so any knees or kinks cannot be used to select the number of clusters.

\section{Results: geometric clustering with DIB\label{sec:Results:-geometric-clustering}}

We ran the DIB as described above on four geometric clustering datasets, varying the smoothing width $s$ (see eqn~\ref{eq:p(x|i)}) and tradeoff parameter $\beta$, and measured for each solution the fraction of spatial information extracted $\tilde{I}\!\left(c;\mathbf{x}\right)=\frac{I\left(c;\mathbf{x}\right)}{I\left(i;\mathbf{x}\right)}$\footnote{Note that $I\!\left(i;\mathbf{x}\right)$ is an upper bound on $I\!\left(c;\mathbf{x}\right)$ due to the data processing inequality,\citep{cover2006infotheory} so $\tilde{I}\!\left(c;\mathbf{x}\right)$ is indeed the fraction of potential geometric information extracted from the smoothed $P\!\left(i,\mathbf{x}\right)$.} and the number of clusters used $n_{c}$ , as well as the kink angle $\theta$. We iterated the DIB equations above just as in \citet{strouse2017dib} with one difference. Iterating greedily from some initialization can lead to local minima (the DIB optimization problem is non-convex). To help overcome suboptimal solutions, upon convergence, we checked whether merging any two clusters would improve the value $L$ of the cost functional in eqn~\ref{eq:DIB_cost}. If so, we chose the merging with the highest such reduction, and began the iterative equations again. We repeated this procedure until the algorithm converged and no merging reduced the value of $L$. We found that these ``non-local'' steps worked well in combination with the greedy ``local'' improvements of the DIB iterative equations. While not essential to the function of DIB, this improvement in performance produced cleaner information curves with less ``noise'' caused by convergence to local minima. Similar to \citet{strouse2017dib}, the automated search over $\beta$ began with an initial set of values, and then iteratively inserted more values where there were large jumps in $H\!\left(c\right)$, $I\!\left(c;\mathbf{x}\right)$, or the number of clusters used, or where the largest value of $\beta$ did not lead to a clustering solution capturing nearly all of the available geometric information (that is, with $I\!\left(c;\mathbf{x}\right)\approx I\!\left(i;\mathbf{x}\right)$. For more details, see our code repository at \href{https://github.com/djstrouse/information-bottleneck}{https://github.com/djstrouse/information-bottleneck}.

\begin{figure*}
\begin{center}
\makebox[\linewidth]{\includegraphics[scale=0.47]{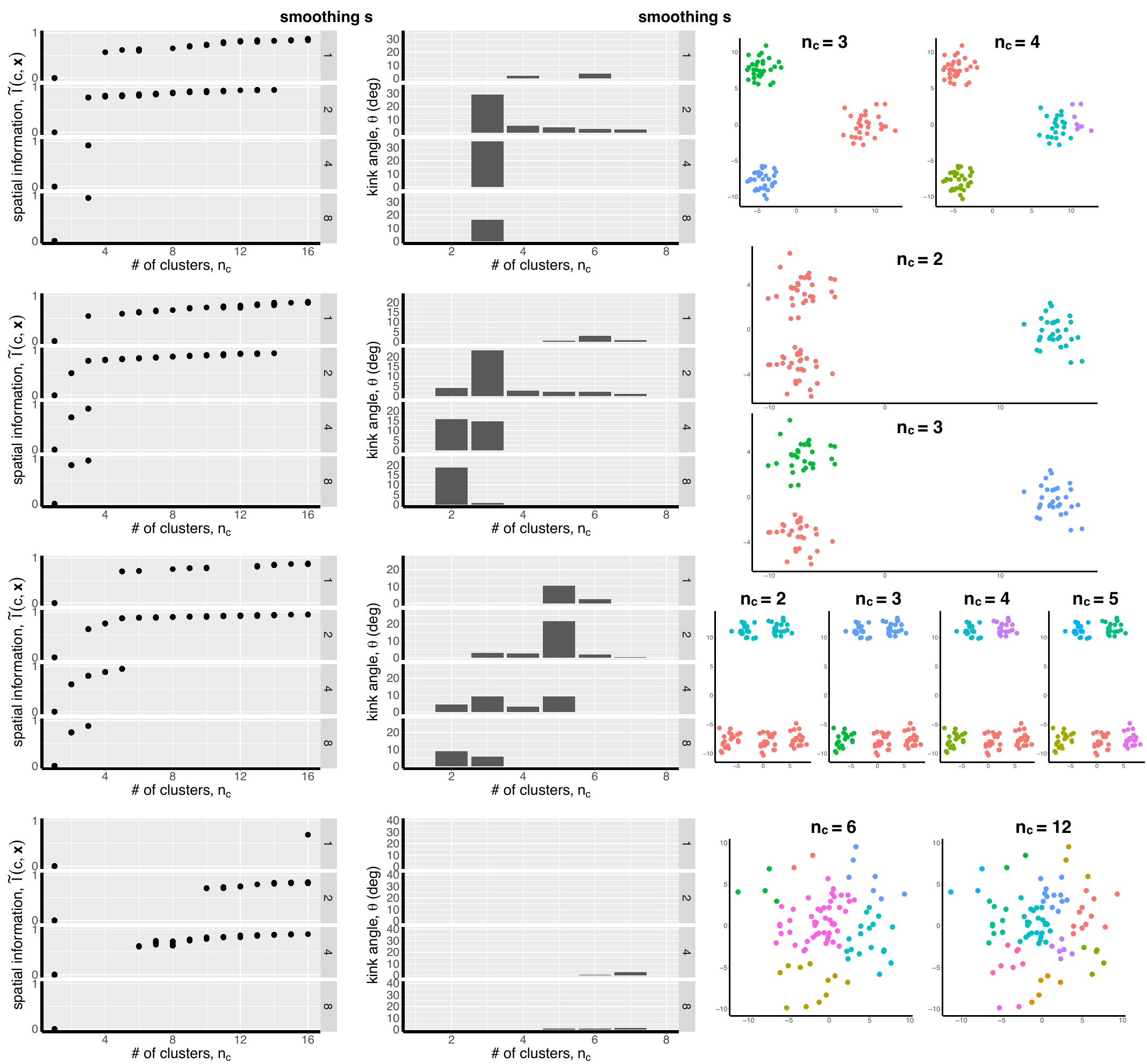}}
\end{center}
\caption{\textbf{Results: model selection and clustering with DIB.}
\label{fig:clusters}
Results for four datasets. Each row represents a different dataset. \emph{Left column}: fraction of spatial information extracted, $\tilde{I}\!\left(c;\mathbf{x}\right)=\frac{I\!\left(c;\mathbf{x}\right)}{I\!\left(i;\mathbf{x}\right)}$, versus number of clusters used, $n_{c}$, across a variety of smoothing scales, $s$. \emph{Center column}: kink angle $\theta$ (of the $I\!\left(c;\mathbf{x}\right)$ vs $H\!\left(c\right)$ curve) versus number of clusters used, $n_{c}$, across a variety of smoothing scales, $s$. \emph{Right column}: example resulting clusters.}
\end{figure*}

Results are shown in figure~\ref{fig:clusters}. Each large row represents a different dataset. The left column shows fractional spatial information $\tilde{I}\!\left(c;\mathbf{x}\right)$ versus number of clusters used $n_{c}$,\footnote{Note that this is \emph{not} the same as the $n_c$ in eqns~\ref{eqn:q(c)}~and~\ref{eqn:q(x|c)}, which was the number of data points assigned to a particular cluster $c$. Here we are using it to denote the number of clusters with at least one data point assigned to it.} stacked by smoothing width $s$.\footnote{Note that this is \emph{not} the same as the information plane curve from figure~\ref{fig:kink}. While the $y$-axes are the same (up to the normalization), the $x$-axes are different.} The center column shows the kink angle $\theta$ for each cluster number $n_{c}$, again stacked by smoothing width $s$. Finally, the right column shows example solutions.

In general, note that as we increase $\beta$, we move right along the plots in the left column, that is towards higher number of clusters $n_{c}$ and more spatial information $\tilde{I}\!\left(c;\mathbf{x}\right)$. Not all values of $n_{c}$ are present because while varying the implicit parameter $\beta$, DIB will not necessarily ``choose'' to use all possible cluster numbers. For example, for small smoothing width $s$, most points won't have enough overlap in $p\!\left(\mathbf{x}\mid i\right)$ with their neighbors to support solutions with few clusters, and for large smoothing width $s$, local spatial information is thrown out and \emph{only} solutions with few clusters are possible. More interestingly, DIB may retain or drop solutions based on how well they match the structure of the data, as we will discuss for each dataset below. Additionally, solutions that match well the structure in the data (for example, ones with $n_{c}$ matched to the generative parameters) tend to be especially robust to $\beta$, that is they have a large kink angle $\theta$. Thus, $\theta$ can be used to perform model selection. For datasets with structure at multiple scales, the kink angle $\theta$ will select different solutions for different values of the smoothing width $s$. This allows us to investigate structure in a dataset at a particular scale of our choosing. We now turn to the individual datasets.

The first dataset (top row) consists of 3 equally spaced, equally sampled symmetric gaussian clusters (see solutions in right column). We see that the 3-cluster solution stands out in several ways. First, it is robust to spatial scale $s$. Second, the 3-cluster solution extract nearly all of the available spatial information; solutions with $n_{c}\geq4$ extract little extra $\tilde{I}\!\left(c;\mathbf{x}\right)$. Third and perhaps most salient, the 3-cluster solution has by far the largest value of kink angle $\theta$ across a wide range of smoothing scales. In the right column, we show examples of 3 and 4-cluster solutions. Note that while all 3-cluster solutions look exactly like this one, the 4-cluster solutions vary in how they chop one true cluster into two.

The second dataset (second row) consists of 3 more equally sampled symmetric gaussian clusters, but this time not equally spaced; two are much closer to one another than the third. This is a dataset with multiple scales present, thus we should expect that the number of clusters picked out by any model selection procedure, e.g. kink angle, should depend on the spatial scale of interest. Indeed, we see that to be true. The 3-cluster solution is present for all smoothing widths shown, but is only selected out as the best solution by kink angle for intermediate smoothing widths ($s=2$). For large smoothing widths] ($s=8$), we see that the 2-cluster solution is chosen as best. For smoothing widths in between ($s=4$), the 2 and 3-cluster solutions are roughly equally valid. In terms of spatial information, the 2 and 3-cluster solutions are also prominent, with both transitions from $n_{c}=1\rightarrow2$ and $n_{c}=2\rightarrow3$ providing significant improvement in $\tilde{I}\!\left(c;\mathbf{x}\right)$ (but little improvement for more fine-grained clusterings).

The third dataset (third row) features even more multi-scale structure, with 5 symmetric, equally sampled gaussians, again with unequal spacing. Sensible solutions exist for $n_{c}=2-5$, and this can be seen by the more gradual rise of the fractional spatial information $\tilde{I}\!\left(c;\mathbf{x}\right)$ with $n_{c}$ in that regime. We also again see a transition in the model selection by $\theta$ from the 5-cluster solution at small smoothing widths ($s=1,2$) and the 2-cluster solution at larger smoothing widths ($s=8$), with intermediate $n_{c}$ favoring those and intermediate solutions. Example clusters for $n_{c}=2-5$ are shown at right.

Finally, we wanted to ensure that DIB and our model selection procedure would not hallucinate structure where there is none, so we applied it to a single gaussian blob, with the hope that no solution with $n_{c}>1$ would stand out and prove robust to $\beta$. As can be seen in the fourth row of figure~\ref{fig:clusters}, that is indeed true. No solution at any smoothing width had particularly high kink angle $\theta$, and no solution remained at the ``knee'' of the $\tilde{I}\!\left(c;\mathbf{x}\right)$ versus $n_{c}$ curve across a wide range of smoothing widths.

Overall, these results suggest that DIB on smoothed data is able to recover generative geometric structure at multiple scales, using built-in model selection procedures based on identifying robust, spatially informative solutions.

\section{Relationship between (D)IB and GMMs \& $k$-means}
\label{sec:GMMs}

It is natural to wonder how the algorithm we introduce here, clustering with DIB, relates to classic approaches to clustering, including GMMs and $k$-means. We now establish the following equivalence: when the smoothing scale $s$ is small, $\beta=1$, and $q\!\left(\mathbf{x}\mid c\right)$ is approximated as a gaussian $r\!\left(\mathbf{x}\mid c\right)$ whose parameters are chosen to minimize $\text{KL}\!\left[p\!\left(\mathbf{x}\mid c\right)\mid r\!\left(\mathbf{x}\mid c\right)\right]$, DIB and IB correspond to EM-fitting of a GMM with hard and soft assignments, respectively. When $s$ is small and $r\!\left(\mathbf{x}\mid c\right)$ is chosen to be an isotropic gaussian with fixed variance across clusters, DIB and IB correspond to hard and soft k-means, respectively, with a logarithmic ''cluster size bonus'' weighted by $\beta^{-1}$. In the $\beta\rightarrow\infty$ limit, the effect of the cluster size bonus vanishes and the correspondence with hard and soft k-means is exact. Thus, clustering with (D)IB can be viewed as a generalization of these approaches.

We begin by establishing the correspondence between DIB and the E-step of fitting a GMM. Consider the KL divergence $\text{KL}\!\left[p\!\left(\mathbf{x}\mid i\right)\mid q\!\left(\mathbf{x}\mid c\right)\right]$ that (D)IB uses to cluster data points. When the smoothing scale $s$ is chosen to be small relative to the scale of $q\!\left(\mathbf{x}\mid c\right)$, then we have
\begin{align}
\text{KL}\!\left[p\!\left(\mathbf{x}\mid i\right)\mid q\!\left(\mathbf{x}\mid c\right)\right]&=-\int p\!\left(\mathbf{x}\mid i\right)\log q\!\left(\mathbf{x}\mid c\right)d\mathbf{x}-H\!\left[p\!\left(\mathbf{x}\mid i\right)\right]\\
&\approx-\log q\!\left(\mathbf{x}_{i}\mid c\right)\int p\!\left(\mathbf{x}\mid i\right)d\mathbf{x}-H\!\left[p\!\left(\mathbf{x}\mid i\right)\right]\\
&=-\log q\!\left(\mathbf{x}_{i}\mid c\right)-H\!\left[p\!\left(\mathbf{x}\mid i\right)\right],	
\end{align}
where we have used the assumption about the scale of $s$ in moving from the first to second line. Since $H\!\left[p\!\left(\mathbf{x}\mid i\right)\right]$ is independent of the cluster assignments, minimizing $\text{KL}\!\left[p\!\left(\mathbf{x}\mid i\right)\mid q\!\left(\mathbf{x}\mid c\right)\right]$ with respect to the cluster assignments is then equivalent to maximizing $\log q\!\left(\mathbf{x}_{i}\mid c\right)$, that is choosing a maximum likelihood assignment of points to clusters. Thus eqn~\ref{eqn:c*} becomes
\begin{align}
c^{*}\!\left(i\right)&=\text{\ensuremath{\underset{c}{\operatorname{argmax}}}}\,\,\log p\!\left(c\right)+\beta\log p\!\left(\mathbf{x}_{i}\mid c\right)\\
&=\text{\ensuremath{\underset{c}{\operatorname{argmax}}}}\,\,\log p\!\left(c\right)^{1/\beta}+\log p\!\left(\mathbf{x}_{i}\mid c\right).\label{eqn:betaMAP}
\end{align}
For $\beta=1$, the two log probabilities combine and lead to a maximum a posteriori (MAP) assignment of points to clusters
\begin{align}
c^{*}\!\left(i\right)&=\text{\ensuremath{\underset{c}{\operatorname{argmax}}}}\,\,\log p\!\left(\mathbf{x}_{i},c\right)=\text{\ensuremath{\underset{c}{\operatorname{argmax}}}}\,\,p\!\left(c\mid \mathbf{x}_{i}\right).\label{eqn:MAP}
\end{align}
For $1 < \beta < \infty$, the effect of $\beta$ is to ''soften'' the prior $p\!\left(c\right)$ (eqn~\ref{eqn:betaMAP}), leading to less aggressive cluster consolidation.

Of course, if we use the exact $q\!\left(\mathbf{x}\mid c\right)$ defined in eqn~\ref{eqn:q(x|c)}, then the scales of $p\!\left(\mathbf{x}\mid i\right)$ and $q\!\left(\mathbf{x}\mid c\right)$ are similar, and so our assumption in this section is not valid. In order for it to be valid, we need to replace the exact $q\!\left(\mathbf{x}\mid c\right)$ with an assumed parametric form that leads to further smoothing.

If we choose to replace $q\!\left(\mathbf{x}\mid c\right)$ with a gaussian approximation $r\!\left(\mathbf{x}\mid c\right)=\mathcal{N}\!\left(\mathbf{x}\mid\mu_{c},\Sigma_{c}\right)$, then eqn~\ref{eqn:MAP} corresponds to the E-step in EM fitting of a GMM \citep{bishop2006prml}. Note that ideally we would like for it to be true that $\text{KL}\!\left[p\!\left(\mathbf{x}\mid i\right)\mid r\!\left(\mathbf{x}\mid c\right)\right]\geq \text{KL}\!\left[p\!\left(\mathbf{x}\mid i\right)\mid q\!\left(\mathbf{x}\mid c\right)\right]$ so that the replacement of $q\!\left(\mathbf{x}\mid c\right)$ by $r\!\left(\mathbf{x}\mid c\right)$ leads to us maximizing a lower bound on our original objective (i.e. that which is maximized in eqn~\ref{eqn:c*}), however this is not generically true and $\text{KL}\!\left[p\!\left(\mathbf{x}\mid i\right)\mid r\!\left(\mathbf{x}\mid c\right)\right]$ might be smaller or larger than $\text{KL}\!\left[p\!\left(\mathbf{x}\mid i\right)\mid q\!\left(\mathbf{x}\mid c\right)\right]$.

The results in this section are only valid for a ''small'' smoothing scale $s$, so let us now understand what that means in the particular case of gaussian $r\!\left(\mathbf{x}\mid c\right)$. Consider the KL divergence in the assignment step (eqn~\ref{eqn:c*}), which in this case has a simple expression
\begin{align}
\text{KL}\!\left[p\!\left(\mathbf{x}\mid i\right)\mid r\!\left(\mathbf{x}\mid c\right)\right]&\propto \frac{s^{2}}{\tr\!\left(\Sigma_{c}\right)}+\left(\mu_{c}-\mathbf{x}_{i}\right)^{T}\Sigma_{c}^{-1}\left(\mu_{c}-\mathbf{x}_{i}\right)+\log\det\Sigma_{c}+k,
\end{align}
where $k$ denotes terms not dependent on the assignment of points to clusters, and thus irrelevant for the objective. Compare to the maximum likelihood objective. The negative log likelihood of $\mathbf{x}_{i}$ under $r\!\left(\mathbf{x}\mid c\right)$ is
\begin{align}
	-\log \mathcal{N}\!\left( \mathbf{x}_{i} \mid \mu_{c}, \Sigma_{c} \right) & \propto \left( \mu_{c} - \mathbf{x}_{i} \right)^{T} \Sigma_{c}^{-1} \left( \mu_{c} - \mathbf{x}_{i} \right) + \log \det\!\left(\Sigma_{c}\right)+k,\label{eqn:nll}
\end{align}
where $k$ again denotes terms independent of the assignment of points to clusters, and thus ignorable. Note that when $s^{2}\ll\tr\!\left(\Sigma_{c}\right)$, the last two equations are the same, and thus the DIB cluster assignments correspond to maximum likelihood assignments. Thus, ''small'' $s$ means $s^{2}\ll\tr\!\left(\Sigma_{c}\right)$ in this case. Of course, we don't know $\tr\!\left(\Sigma_{c}\right)$ until after we cluster our data, but it is set by the natural length scales in the data, so we can take it to mean that $s$ needs to be small compared to those.

That establishes the correspondence for the E-step of EM fitting of a GMM, but what about the M-step? Note that we haven't yet specified how to fit the approximation $r\!\left(\mathbf{x}\mid c\right)\approx q\!\left(\mathbf{x}\mid c\right)$. One reasonable way that appears often in the variational inference literature (e.g. \citet{kingma2014vae}) is to choose the parameters of $r\!\left(\mathbf{x}\mid c\right)$ ($\mu_{c}$ and $\Sigma_{c}$) that minimize $\text{KL}\!\left[p\!\left(\mathbf{x}\mid c\right)\mid r\!\left(\mathbf{x}\mid c\right)\right]$. We choose this direction of the KL divergence because it encourages a ''mean-seeking'' approximation of $p\!\left(\mathbf{x}\mid c\right)$ that tries better to approximate the full distribution than the other, ''mode-seeking'' direction. While this is again a generally intractable KL divergence between a mixture of gaussians and a gaussian, fortunately in the $s^{2}\ll\tr\!\left(\Sigma_{c}\right)$ limit that we consider, it simplifies to
\begin{align}
\text{KL}\!\left[p\!\left(\mathbf{x}\mid c\right)\mid r\!\left(\mathbf{x}\mid c\right)\right]&=-\int p\!\left(\mathbf{x}\mid c\right)\log r\!\left(\mathbf{x}\mid c\right)d\mathbf{x}-H\!\left[p\!\left(\mathbf{x}\mid c\right)\right]\label{eqn:KL[p|r]}\\
&=-\frac{1}{n_{c}}\int\sum_{i\in S_{c}}\mathcal{N}\!\left(\mathbf{x};\mathbf{x}_{i},s^{2}\right)\log r\!\left(\mathbf{x}\mid c\right)d\mathbf{x}-H\!\left[p\!\left(\mathbf{x}\mid c\right)\right]\\
&\approx-\frac{1}{n_{c}}\sum_{i\in S_{c}}\log r\!\left(\mathbf{x}_{i}\mid c\right)\int\mathcal{N}\!\left(\mathbf{x};\mathbf{x}_{i},s^{2}\right)d\mathbf{x}-H\!\left[p\!\left(\mathbf{x}\mid c\right)\right]\\
&=-\frac{1}{n_{c}}\sum_{i\in S_{c}}\log r\!\left(\mathbf{x}_{i}\mid c\right)-H\!\left[p\!\left(\mathbf{x}\mid c\right)\right],\label{eqn:Mstep}
\end{align}
where we move from the second to third line using the small $s$ approximation (so that $r\!\left(\mathbf{x}\mid c\right)\approx r\!\left(\mathbf{x}_{i}\mid c\right)$ in the region of $\mathbf{x}$ where the bulk of $\mathcal{N}\!\left(\mathbf{x};\mathbf{x}_{i},s^{2}\right)$ is). Minimizing eqn~\ref{eqn:Mstep} with respect to $\mu_{c}$ and $\Sigma_{c}$ again corresponds to maximum likelihood assignments, this time of the model parameters rather than cluster assignments. This corresponds to the M-step of EM fitting of a GMM \citep{bishop2006prml}.

Thus, for $\beta=1$, small $s$, and a gaussian approximation of $p\!\left(\mathbf{x}\mid c\right)$ (with parameters chosen to minimize the KL divergence in eqn~\ref{eqn:KL[p|r]}), clustering with DIB is equivalent to EM fitting of a GMM with hard assignments (of data points to clusters). For $\beta > 1$, the effect of the cluster prior $p\!\left(c\right)$ is muted; that is, it is replaced with $p\!\left(c\right)^{1/\beta}$.

If we set all cluster conditional approximations to have the same isotropic covariance $\Sigma_{c}=\text{diag}\!\left(\sigma^{2}\right)$, then $c^{*}\left(i\right)$ becomes (plugging eqn~\ref{eqn:nll} into eqn~\ref{eqn:betaMAP})
\begin{align}
	c^{*}\!\left(i\right)&=\text{\ensuremath{\underset{c}{\operatorname{argmax}}}}\,\, \frac{\sigma^2}{\beta}\log p\!\left(c\right)-\left\Vert \mathbf{x}_{i}-\mu_{c}\right\Vert ^{2}\\
	&=\text{\ensuremath{\underset{c}{\operatorname{argmax}}}}\,\, \frac{\sigma^2}{\beta}\log n_c-\left\Vert \mathbf{x}_{i}-\mu_{c}\right\Vert ^{2},
\end{align}
which corresponds to (hard) $k$-means with a cluster size bonus $\log n_{c}$ (where $n_{c}\equiv \left| S_c\right| $ is the number of points assigned to cluster $c$, as introduced in section~\ref{sec:clustering-with-DIB}). In the $\beta\rightarrow\infty$ limit, the $\log n_{c}$ term can be ignored and the correspondence with (hard) $k$-means is exact.

To see the correspondence between GMMs/$k$-means and IB, consider that IB can be viewed as DIB with the hard max replaced by a soft max (see eqn~\ref{eqn:IBenc}.) Thus, the same correspondences we drew between DIB and GMMs/$k$-means with \emph{hard} assignments hold for IB and GMMs/$k$-means with \emph{soft} assignments.

The correspondence between clustering with (D)IB and GMMs yields new interpretations of both. From this perspective, clustering with (D)IB can be viewed as a generalization of GMMs that 1) uses a more flexible, nonparametric representation of the clusters, 2) includes an extra parameter $\beta$ for controlling the tradeoff between the prior and likelihood, and 3) includes an extra parameter $s$ for setting the length scale of interest. In the other direction, GMMs can be viewed as mapping data points to cluster labels that maximally preserve spatial information.

This is not the first correspondence between IB in a particular setting and another probabilistic model. In the discrete setting, IB has been shown to be related to EM fitting of a multinomial mixture model \citep{slonim2003MLandIB}. In the time series setting (where $X=x_t$ and $Y=x_{t+1}$), IB is related to canonical correlation analysis \citep{creutzig2009pastfuture}, and therefore linear gaussian models \citep{bach2006cca} and slow feature analysis \citep{turner2007MLandSFA}. Under a variational approximation, IB applied to unsupervised learning is related to a variational autoencoder (VAE) \citep{alemi2017vib,higgins2017betavae,kingma2014vae}.

\section{Discussion}

Here, we have shown how to use the formalism of the information bottleneck to perform geometric clustering. A previous paper \citep{still2004geoIB} claimed to contribute similarly, however for the reasons discussed in sections~\ref{sec:clustering-with-DIB}~and~\ref{sec:Appendix:-errors}, their approach contained fundamental flaws. We amend and improve upon that paper in four ways. First, we show to fix the errors they made in their problem setup (with the data preparation). Second, we argue for using DIB over IB in this setting for its preference for using as few clusters as it can. Third, we introduce a novel form of model selection for the number of clusters based on discontinuities (or ``kinks'') in the slope of the DIB curve, which indicate solutions that are robust across the DIB tradeoff parameter $\beta$. We show that this information-based model selection criterion allows us to correctly recover generative structure in the data at multiple spatial scales. Finally, we establish the correct correspondence between clustering with (D)IB and $k$-means/GMMs, thus providing both a generalization and information-theoretic interpretation of these classic approaches.

We have introduced one way of doing geometric clustering with the information bottleneck, but we think it opens avenues for other ways as well. First, the uniform smoothing we perform above could be generalized in a number of ways to better exploit local geometry and better estimate the ``true'' generative distribution of the data. For example, one could do gaussian smoothing with mean centered on each data point but the covariance estimated by the sample covariance of neighboring data points around that mean. Indeed, our early experiments with this alternative suggest it may be useful for certain datasets. Second, while choosing spatial location as the relevant variable for DIB to preserve information about seems to be the obvious first choice to investigate, other options might prove interesting. For example, preserving information about the identity of neighbors, if carefully formulated, might make fewer implicit assumptions about the shape of the generative distribution, and enable the extension of our approach to a wider range of datasets.

Scaling the approach introduced here to higher-dimensional datasets is non-trivial because the tabular representation used in the original IB \citep{tishby1999ib} and DIB \citep{strouse2017dib} algorithms leads to an exponential scaling with the number of dimensions. Recently, however, \citet{alemi2017vib} introduced a variational version of IB, in which one parameterizes the encoder $q\!\left(t\mid x\right)$ (and ``decoder'' $q\!\left(y\mid t\right)$) with a function approximator, e.g. a deep neural network. This has the advantage of allowing scaling to much larger datasets. Moreover, the choice of parameterization often implies a smoothness constraint on the data, relieving the problem encountered above of needing to smooth the data. It would be interesting to develop a variational version of DIB, which could then be used to perform information-theoretic clustering as we have done here, but on larger problems and perhaps with no need for data smoothing.

\section{Appendix: errors in \citet{still2004geoIB}\label{sec:Appendix:-errors}}

A previous attempt was made to draw a connection between IB and $k$-means \citep{still2004geoIB}. Even before reviewing the algebraic errors that lead their result to break down, there are two intuitive reasons why such a claim is unlikely to be true. First, IB is a soft clustering algorithm, and $k$-means is a hard clustering algorithm. Second, the authors made the choice not to smooth the data and to set $p\!\left(\mathbf{x}\mid i\right)=\delta_{\mathbf{x}\mathbf{x}_{i}}$. As discussed in section~\ref{sec:clustering-with-DIB}, (D)IB clusters data points based on these conditionals, and delta functions trivially only overlap when they are identical.

The primary algebraic mistake appears just after eqn~14, in the claim that $p_{n}\!\left(\mathbf{x}\mid c\right)\propto p_{n-1}\!\left(\mathbf{x}\mid c\right)^{1/\lambda}$. Combining the previous two claims in that proof, we obtain:
\begin{align}
p_{n}\!\left(\mathbf{x}\mid c\right) & =\frac{1}{N}\sum_{i}\frac{\delta_{\mathbf{x}\mathbf{x}_{i}}}{Z_{n}\!\left(i,\lambda\right)}p_{n-1}\!\left(\mathbf{x}_{i}\mid c\right)^{1/\lambda}.
\end{align}

Certainly, this does not imply that $p_{n}\!\left(\mathbf{x}\mid c\right)\propto p_{n-1}\!\left(\mathbf{x}\mid c\right)^{1/\lambda}$ everywhere, because of the $\delta_{\mathbf{x}\mathbf{x}_{i}}$ factor which picks out only a finite number of points.

One might wonder why with these mistakes, the authors still obtain an algorithm that looks and performs like $k$-means. The reason is because their sequence of mistakes leads to the result in eqn~15 that effectively \emph{assumes} that IB has access to geometric information it should not, namely the cluster centers at step $n$. Since these are exactly what $k$-means uses to assign points to clusters, it is not surprising that the behavior then resembles $k$-means.

\section{Acknowledgements}

We would like to thank Léon Bottou and Arthur Szlam for useful discussions. We would also like to acknowledge financial support from the NSF, through the Center for the Physics of Biological Function (PHY-1734030) (Schwab), the Simons Foundation (Schwab), and the Hertz Foundation (Strouse).

\bibliographystyle{apa}
\bibliography{arxiv.bib}

\end{document}